\documentclass[letterpaper, 10 pt, conference]{ieeeconf}  

\let\labelindent\relax

\IEEEoverridecommandlockouts                              

\usepackage[utf8]{inputenc}
\usepackage{comment}
\usepackage{graphicx}
\usepackage{subcaption}
\usepackage{cleveref}
\usepackage{enumitem}
\usepackage{booktabs}
\usepackage{gensymb}
\usepackage[font=footnotesize]{caption}
\usepackage{cite}
\usepackage{todonotes}
\usepackage{tabularx}

\usepackage{titlesec}
\titlespacing{\section}{5pt}{*0.7}{*0.7}
\titlespacing{\subsection}{10pt}{*0.6}{*0.6}
\usepackage{enumitem}
\setlist[itemize]{noitemsep, topsep=0pt}
\setlist[enumerate]{noitemsep, topsep=0pt}

\title{\LARGE \bf Seeing Thru Walls: Visualizing Mobile Robots in Augmented Reality}
\author{Morris Gu, Akansel Cosgun, Wesley P. Chan, Tom Drummond and Elizabeth Croft \\
Monash University, Australia}

\begin{document}
\maketitle

\begin{abstract}
    We present an approach for visualizing mobile robots through an Augmented Reality headset when there is no line-of-sight visibility between the robot and the human. Three elements are visualized in Augmented Reality: 1) Robot's 3D model to indicate its position, 2) An arrow emanating from the robot to indicate its planned movement direction, and 3) A 2D grid to represent the ground plane. We conduct a user study with 18 participants, in which each participant are asked to retrieve objects, one at a time, from stations at the two sides of a T-junction at the end of a hallway where a mobile robot is roaming. The results show that visualizations improved the perceived safety and efficiency of the task and led to participants being more comfortable with the robot within their personal spaces. Furthermore, visualizing the motion intent in addition to the robot model was found to be more effective than visualizing the robot model alone. The proposed system can improve the safety of automated warehouses by increasing the visibility and predictability of robots.
\end{abstract}
  
\section{Introduction}
At Amazon fulfillment centers, it was found that warehouses which utilised robots had a rate of serious injury greater than 50\% higher when compared to warehouses without robots.~\cite{Amazon}. Worker safety is an important consideration for the large-scale deployment of mobile robots in environments where humans and robots co-exist~\cite{cosgun2013accuracy,lasota2017survey}. This includes both physical safety, where the robot does not cause physical injury to humans, and mental safety, where the robot does not cause fear or shock~\cite{alami2006safe}. Robots that move in predictable ways arguably improve the safety of the human-robot interaction in both of those aspects. Methods for generating predictable robot motion has previously used gaze~\cite{fischer_between_2016,zheng2015impacts}, light projection~\cite{fernandez_passive_2018}, plan visualization~\cite{newbury_visualizing_2021} or implicit robot motions~\cite{cosgun2016anticipatory,dragan2013legibility}, however, these approaches assume that the robot is visible to the human. When the robot is not visible to the humans in the environment, it would not be possible to know where the robot is and where it is moving towards, or even if there is a robot in the occluded areas. When the robot is in the occluded areas and abruptly appears to humans, it can surprise and scare them, especially when the robot is close to them.~\cite{sisbot2007human}. This is a common occurrence, for instance, around corners or T-junctions at hallways~\cite{chung2009safe}. Past research tried to address this problem as part of the robot path planning by the introduction of a `visibility' cost to discourage the robot from entering into the areas that are invisible to humans~\cite{morales2015including,sisbot2007human}. However, these planners do not address the user's perception of safety. Hence, a new approach is desirable.

In this paper, we propose a visualization solution from the user's perspective, giving the user the ability to see robots through walls rather penalizing the robot's path planning. We propose visualizing three kinds of information through an Augmented Reality (AR) headset: 
\begin{enumerate}
    \item The robot's 3D model at the robot's current pose.
    \item An arrow, emanating from the robot, to indicate the short-term movement direction of the robot.
    \item A 2D grid to represent the ground plane.
\end{enumerate}

We consider a scenario where a mobile robot and a human have separate tasks but will need to cross paths multiple times at a T-junction. We assume that both the robot and the AR headset are localized in a shared global map, therefore the pose of the robot can be computed from the headset's view. The proposed method is evaluated through a user study, in which we measure the effect of visualizing the robot model, as well as its motion intent. A subject is assigned an auxiliary task in a hallway where a robot is initially occluded behind a T-junction. The task consists of the users walking over to one of the two object stations and picking the desired object from around the corner, where a mobile robot may be navigating. This scenario is chosen to mimic an order-picking task where encountering a mobile robot is possible, for example in a mixed-automation warehouse or assembly floor.

\begin{figure}[t!]
    \vspace{0.25cm}
    \centering
    \includegraphics[trim= 0 0 68 30,clip,width=0.65\linewidth]{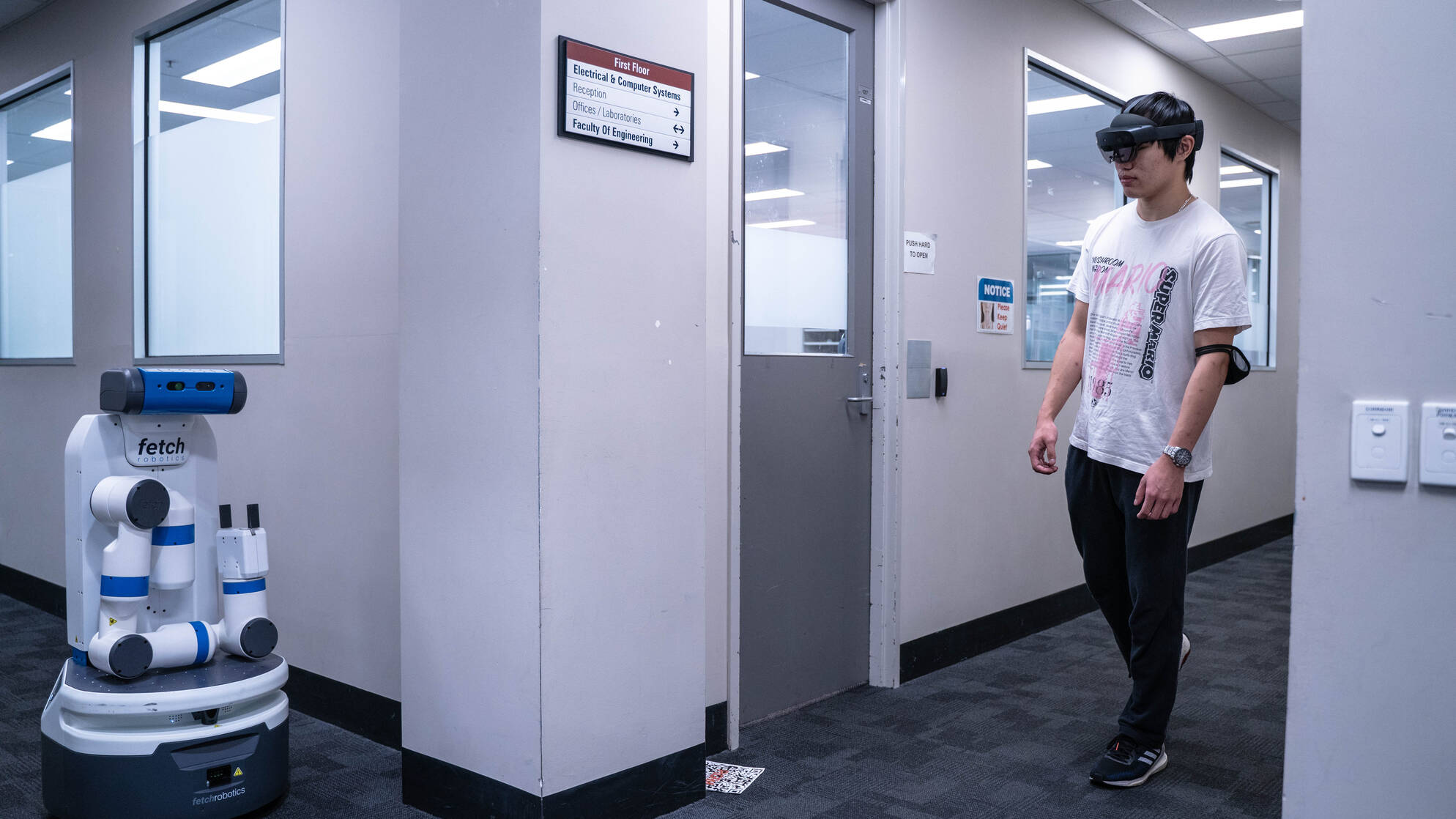}\hspace{0cm}
    \vspace{0.1cm}
    \includegraphics[trim= 0 0 230 0,clip, width=0.33\linewidth]{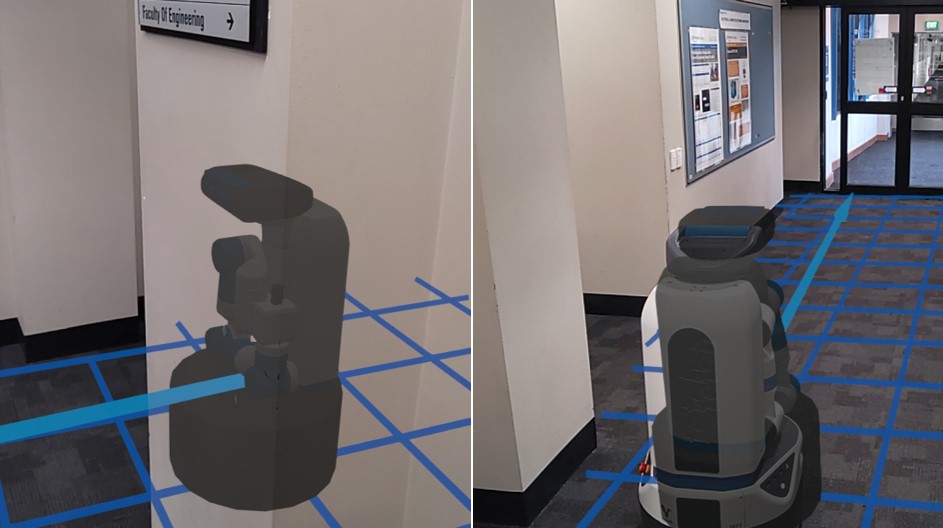}
    \caption{Left: The mobile robot is not visible to the user at a T-Junction. Right: A representation of what the user sees from the AR headset. The robot model is visualized, enabling the user to see the robot through the wall. The ground is shown as a grid to improve depth perception.} 
    \label{fig:intro}
\end{figure}

The contributions of this paper are two-fold:
\begin{itemize}
    \item To our knowledge, the first use of Augmented Reality to visualize a mobile robot's position and its intended future path, even when the robot is not visible to the user
    \item User studies showing the effectiveness of the proposed AR-based interaction design for a scenario that mimics a warehouse picking task.
\end{itemize}

The organization of this paper is as follows. After examining the existing literature on robot intent visualization approaches in Sec.~\ref{sec:RelatedWork}, we explain the details of our visualization technique in Sec.~\ref{sec:approach}. We state our hypotheses in Sec.~\ref{sec:hypotheses} and describe the user study procedure in Sec.~\ref{sec:user_study}. The study results are presented in Sec.~\ref{sec:results}, before concluding and discussing future work in Sec.~\ref{sec:conclusion}.

\section{Related Works}
\label{sec:RelatedWork}

\subsection{Visualizing Occluded Objects in AR}


Lilija~\cite{lilija_augmented_2019} and Gruenefeld~\cite{gruenefeld_locating_2019} both researched applications of occlusion visualization. Lilija~\cite{lilija_augmented_2019} displayed the occluded information to the user using multiple methods and found that users had a preference towards an appropriate point of view and view stability. The \textit{see-through view} presented in their work is the most applicable to the system presented in our paper, with the remaining methods relying on the static scenarios presented by Lilija.

Gruenefeld~\cite{gruenefeld_locating_2019} displayed markers for in-view and/or out-of-view visualizations, finding that each marker type is beneficial individually and combining both leads to the lowest search times. However, using both visualizations introduced visual clutter and distraction which can be a detriment in industrial scenarios where collisions can have serious consequences.

Previous related work by Livingston focused on developing the best depiction of occluded objects~\cite{livingston_resolving_2003}. They showed that a consistent ground plane, and decreasing opacity and intensity as indicators of increasing distance produced results that agree with psychophysical literature. In our work, we adopted a grid-based visualization to indicate distance to users and provide environmental context. This is important due to the small field of view (FOV) of current AR technologies which limits the effectiveness of a consistent ground plane at short distances. 

\subsection{Communicating Robot Intent}
Humans use many nonverbal motion cues to signal motion intention, such as gaze and head movements\cite{huang_using_2015} but these nonverbal cues become difficult in human-robot interaction. Fischer~\cite{fischer_between_2016} studied the effect of robotic gaze and found that users are more comfortable when the robot maintains eye contact with them, compared to gazing at its goal location. Another method,  the use of indicator lights or patterns to indicate motion, is studied by Fernandez~\cite{fernandez_passive_2018}, Szafir~\cite{szafir2015communicating} and May~\cite{may_show_2015}. May compares indicator lights against robotic gaze and found that indicator lights provided greater comfort as users pass the robot. This allowed them to better understand robot motion intent compared to robotic gaze. Szafir~\cite{szafir2015communicating} and Fernandez~\cite{fernandez_passive_2018}, found that light signals also provided more clarity and predictability to robot motion intent.

Baraka~\cite{baraka_enhancing_2016} studied the use of expressive lights on a robot and found that it can be an effective non-verbal communication modality with minimal contextual clues. Similarly, Cha~\cite{cha2016using} found that simple binary sound and light signals are effective at expressing different levels of urgency which can affect the response time of a user. These methods did not provide directional intent but can complement others and, with the aforementioned methods, allow robot intent to be communicated to the user. However, AR or projection methods can provide more clarity to users.

\subsection{Visualizing Robot Intent through Projection and AR}
Recently, projection methods and AR have been studied to provide user a visualization robot intent.
Shrestha~\cite{shrestha_communicating_2018} found that projection of robot motion intent, shown by arrows, is favorable compared to no indication, increasing understanding of robot motion intent and user comfort . Hetherington found that using a projector to display both immediate and future paths is more effective than using flashing lights similar to the methods studied in\cite{fernandez_passive_2018, may_show_2015}. Andersen~\cite{andersen_projecting_2016} compared projector methods against text description and monitor display methods. They found that the projector method produced less errors and was on average best for user satisfaction and usability.
Walker~\cite{walker_communicating_2018} compared visual indicators for robot motion intent. They found that visualization of navigation points was the most effective, followed by arrows which both indicate the robots path. These increased efficiency and improved subjective understanding of robot motion intent.

\section{Visualizing Robots in Augmented Reality}
\label{sec:approach}

\begin{figure}[h!]
    \centering
    \includegraphics[width = 1\linewidth]{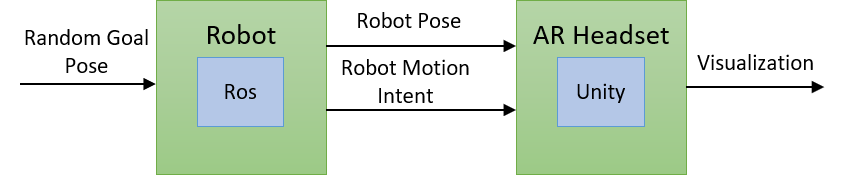}
    \caption{System overview.} 
    \label{fig:system}
\end{figure}

Our system consists of two modules, as shown in Figure \ref{fig:system}, the \textbf{Robot} module which uses Robot Operating System (ROS) and the \textbf{AR Headset} module which uses Unity Engine. The \textbf{Robot} module handles robot navigation by receiving a random robot goal pose shown in Figure \ref{fig:exp_set}. The navigation behavior will be further explained in Subsection \ref{subsec:navigation}. The \textbf{Robot} module will send the robot's pose and motion intent to the \textbf{AR Headset} module to be visualized.  

\vspace{-0.5cm}
\subsection{Hardware and Software}


\textbf{Robot}: We use a Fetch Robotics Mobile Manipulator. Only the robot's base is fully utilized as the robot arm is not needed for our work. The robot has a maximum speed of 1.0m/s, an ankle-height SICK 2D LIDAR sensor with a range of 25m and 220 degrees FOV, and a Primesense Carmine RGB-D camera mounted on the head of the robot.

\textbf{AR Headset}: We use A Microsoft HoloLens 2 (HoloLens) which can display virtual objects using see-through holographic lenses. The device provides head position tracking and can localize itself with respect to a fixed reference frame. 

Communication between the devices is achieved through a WiFi router over a local area network. We use ROS for inter-node communication on the robot side, and Unity for the HoloLens side, and ROS\# package for communicating in between the two devices.

\subsection{Localizing AR Headset and the Robot}

The robot and the AR Headset must be co-localized in a shared reference frame in order to be able to compute the visualization objects from the AR Headset's view. 

\textbf{Robot Localization:} We use off-the-shelf packages standard in the ROS Navigation Stack for mapping and localization of the robot: OpenSLAM gmapping and Monte Carlo localization.

\textbf{AR Headset Localization:} We use off-the-shelf localization solution provided by Vuforia~\cite{Vuforia} and HoloLens head tracking. An artificial AR marker is used to initiate the localization, and serves as a reference frame for reporting the 6D pose of the headset. Once the localization is initialized, the AR marker does not need to be in the FOV and the headset continues to localize itself. It is possible that the headset localization drifts or is lost. When this happens, the user can walk up to the the AR marker, look at it, and utter the word ``Calibrate" to re-initialize the localization. During our experiments, the robot never lost its localization, however, the AR headset occasionally lost localization.

\textbf{Co-localization:} The AR marker which is used to initialize headset localization is placed at a fixed location (on the floor in our case), and the pose of the marker with respect to the robot map frame is manually measured and hard-coded before operation. The localization errors for both the robot and the headset is additive, which results in a slight mismatch between the actual robot and its virtual 3D model. The networking time delays also contribute to the localization error, however, users did not comment that these errors significantly effected their perception of the system.

\subsection{Robot Navigation}
\label{subsec:navigation}
In this paper, as depicted in Fig.~\ref{fig:exp_set} and will be explained in detail in Sec.~\ref{subsec:procedure}, the robot navigates between positions where there are no static obstacles, and the only obstacle that can be encountered is the user. Because of the simplicity of the planning task, we use a simple navigation behavior for this work, adopted from our previous work~\cite{newbury2019learning}. Collision is prevented by detecting whether obstacles are within a certain distance in front of the robot. The 2D laser scan is used as an input into the robot navigation module and detects for values within a certain threshold distance in a roughly 100$\degree$ cone in front of the robot. If the values are detected within the threshold, the robot slows down and then reverses away from the obstacle until the distance values are above the threshold. The maximum velocity of the robot is 1 m/s.


\subsection{Visualization in Augmented Reality}
We propose visualizing three elements, to inform the user of the robot's position and motion intent. All objects are rendered with a refresh rate of 60 Hz:
\begin{enumerate}
\item \textbf{Floor Grid:} Grid lines visualize the floor of the environment aiming to enhance the user's depth perception and the sense of distance to the virtual models. The grid lines are pre-processed, generated based on the robot map. The corridor area where the grid is enabled is annotated manually, however, this process is automated by detecting corridors from maps~\cite{bormann2016room}. The orientation of the grid lines is chosen arbitrarily. 
\item \textbf{Robot Model: }The 3D model of the Fetch robot visualized on the robot's real-world pose, as long as it is in the FOV of the AR Headset, regardless of whether it is under occlusion or not. We use a simplified 3D model to reduce the computation required to draw the robot model.The robot model's position is updated at 10 Hz.
\item \textbf{Robot Motion Intent:} An arrow that points to the robot's intended goal starting at the current real-world position of the robot. The arrow covers the whole distance to the intended goal and its position is updated at 10 Hz. For this work, the robot only navigates on linear paths as described in the previous section, therefore the arrow also represents the future path of the robot. How to visualize the future motion intent for arbitrary paths is an open research question~\cite{hetherington_design_2020}.
\end{enumerate}

\section{User Study Design}
\label{sec:user_study}
A user study was conducted to test the effect of robot and motion intent visualization in AR. The methodology is inspired by our previous work on human-robot collaboration~\cite{bansal_supportive_2020} and AR~\cite{newbury_visualizing_2021}.

\subsection{Hypotheses}
\label{sec:hypotheses}

We expect that visualizing the robot through walls would improve the perceived safety and perceived task efficiency of the participants. We also expect that the addition of robot motion intent to the 3D robot model would further improve the perceived safety and perceived task efficiency. We formulate the following hypotheses to test on a user study with a robot: 

\begin{enumerate}[label=\textbf{H\arabic*}]
    \item The proposed AR-based robot visualization method will reduce the perceived risk of collision compared to having no visualizations.
    \item The proposed AR-based robot visualization method will increase the perceived task efficiency compared to having no visualizations.
    \item Visualizing the motion intent in addition to the robot model would reduce the perceived risk of collision.
    \item Visualizing the motion intent in addition to the robot model would increase the perceived task efficiency.
\end{enumerate}

\subsection{Independent Variable}
Visualization mode is the only independent variable in this user study. Three visualization conditions are considered:
\begin{enumerate}
    \item \textbf{No AR:} The user wears the AR Headset but there is no visualization.
    \item \textbf{Robot Only:} The floor grid and the robot model are visualized in AR.
    \item \textbf{Robot+Intent:} The floor grid, the robot model and the robot's motion intent are visualized in AR.
\end{enumerate}

Each participant experienced each condition in the shown order. The order is fixed to introduce a single visualization object (the robot or the motion intent) at a time. We acknowledge this may introduce ordering effects, however, in this study we are interested in measuring the effect of adding in one additional feature.

\subsection{Participant Allocation}
We recruited 18 participants ($13$ male, $5$ female) from within Monash University\footnote[1]{Due to the ongoing COVID-19 pandemic, no external participants could be recruited. This study has been approved by the Monash University Human Research Ethics Committee (Application ID: 27685)}, aged $20-27$ (M = 22.9, SD = 2.04). The participants were not compensated for their time.

\begin{figure}
    \centering
    \vspace{0.25cm}
    \includegraphics[width=1\linewidth]{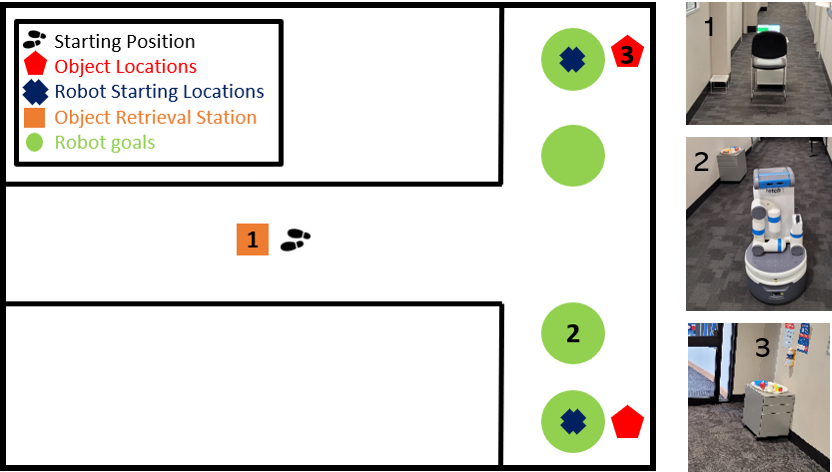}
    \caption{The experimental setup for the object retrieval task with examples}
    \label{fig:exp_set}
\end{figure}

\subsection{Procedure}
\label{subsec:procedure}
\begin{figure}[t]
    \centering
    \includegraphics[width=0.5\linewidth]{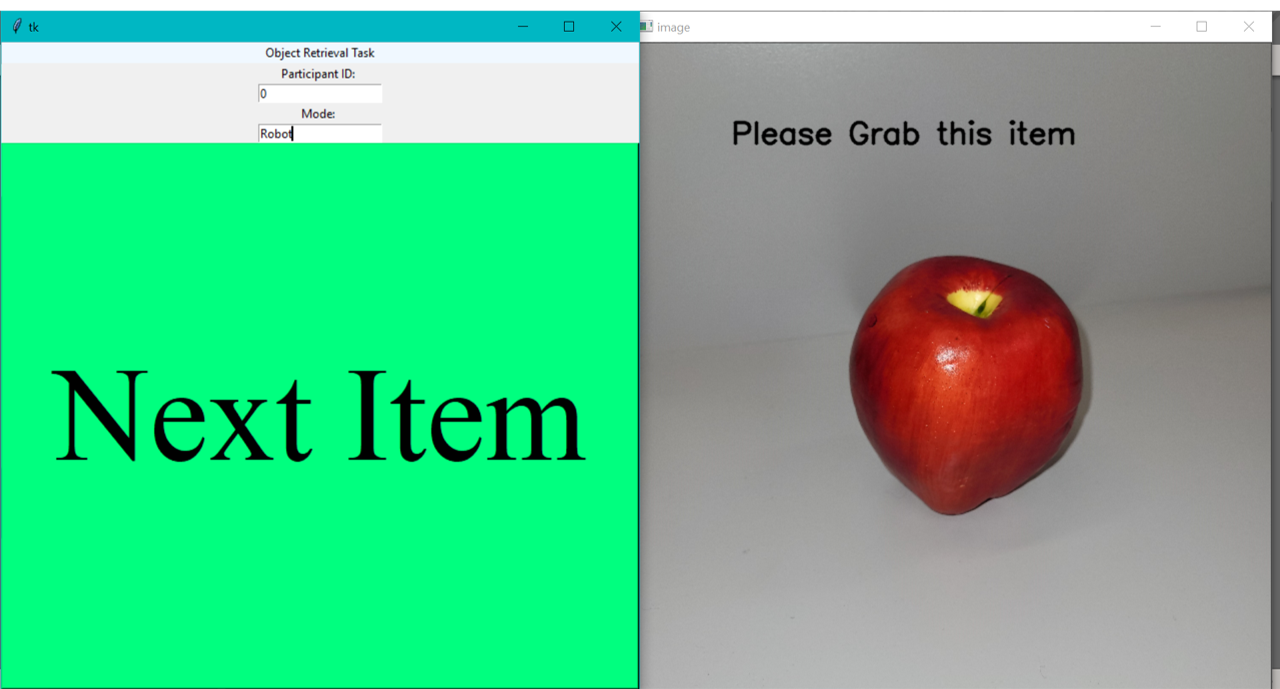}
    \caption{The touch-based user interface shows the participants an image of the next object to retrieve.}
    \label{fig:laptop}
\end{figure}

\begin{figure}[t!]
\centering
\vspace{-0.1cm}
\begin{subfigure}[t]{0.162\linewidth}
    \includegraphics[width=1\linewidth]{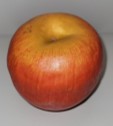}
    \caption{Apple}
    \label{fig:apple}
\end{subfigure}%
\begin{subfigure}[t]{0.162\linewidth}
    \includegraphics[width=1\linewidth]{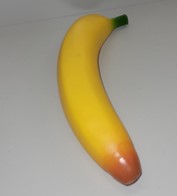}
    \caption{Banana}
    \label{fig:banana}
\end{subfigure}%
\begin{subfigure}[t]{0.162\linewidth}
    \includegraphics[width=1\linewidth]{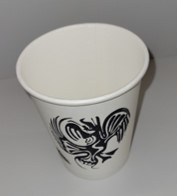}
    \caption{Cup}
    \label{fig:cup}
\end{subfigure}%
\begin{subfigure}[t]{0.162\linewidth}
    \includegraphics[width=1\linewidth]{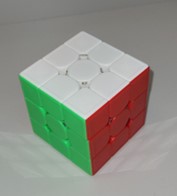}
    \caption{Cube}
    \label{fig:cube}
\end{subfigure}%
\begin{subfigure}[t]{0.162\linewidth}
    \includegraphics[width=1\linewidth]{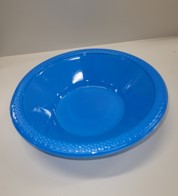}
    \caption{Bowl}
    \label{fig:bowl}
\end{subfigure}%
\begin{subfigure}[t]{0.162\linewidth}
    \includegraphics[width=1\linewidth]{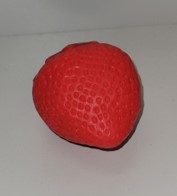}
    \caption{Strawberry}
    \label{fig:strawberry}
\end{subfigure}%
\caption{Set of objects to be retrieved by participants during the experiment.}
\label{fig:objects}
\end{figure}

The experiment took place within a T-junction adjacent to a university lab under the supervision of an experimenter. Participants first read from the explanatory statement and then signed a consent form. The experimenter had participants watch a video to explain the task and read from a script to explain the rest of the experiment.
Participants initially sat at a workstation (location 1 in Fig \ref{fig:exp_set}) and moved between this location and object picking locations shown in Fig \ref{fig:exp_set} to retrieve items. The participants tapped a touchscreen interface shown in Fig \ref{fig:laptop} to find out which object to retrieve next. Users were able to freely choose which object picking location they retrieved objects from, with both stations containing the same 6 unique objects.
Each participant was required to retrieve 5 objects out of 6 total unique objects shown in Fig \ref{fig:objects} per trial, for a total of 15 objects over 3 trials. Before each trial, the robot was located in one of two starting positions shown in Fig \ref{fig:exp_set}. During each trial, the participants wore the HoloLens and active noise cancelling headphones. The headphones played industrial sounds to mimic potential real world contexts and prevent participants from hearing the Fetch robot move. This task was designed to be simple to allow the effect of the visualisation objects to be emphasized. The participants completed up to three types of surveys, a survey after each of the three test conditions, a post-experiment demographic survey, and a post-experiment feedback form. The participants also had the option to engage in a post-experiment interview instead of the post-experiment feedback form. 
The mean duration of each experiment was approximately 50 minutes.

\subsection{Metrics}

\textbf{Subjective Metrics:} After each of the three trials, the participant was asked to answer a number of survey questions. All questions were measured on a 7 point Likert scale. The questions are shown in Table~\ref{tab:questions}. Since \textbf{Q3}-\textbf{Q6} are visualization-specific, participants answered only \textbf{Q1} and \textbf{Q2} after the first trial which is the \textbf{No AR} condition.

\begin{table}[h!]
\vspace{0.25cm}
  \begin{tabular}{|l|}
  
    \toprule
    \textbf{Q1}: I feel collisions with the robot can happen at any time. (R)\\
    \textbf{Q2}: I feel that I completed the task without obstruction. \\
    \textbf{Q3}: The visualization aided my ability to complete task.\\
    \textbf{Q4}: The visualization aided my ability to prevent collisions.\\
    \textbf{Q5}: The visualization is easy to understand.\\
    \textbf{Q6}: The visualization provided useful information.	\\
    \bottomrule
    \end{tabular}
    \caption{Survey questions used in our study. (R) indicates reverse scale. 1 means strongly disagree, 7 means strongly agree}
  \label{tab:questions}
\end{table}
This questionnaire was designed to focus on the novel robot visualisation aspect of the system. At the conclusion of the experiment, the subjects were also asked about what most useful feature of the visualization was. 


\textbf{Objective Metrics:} We use two objective metrics: task completion time and social comfort. We define task completion time to be the average retrieval time of an object, which is measured as the time between successive laptop interface touches in seconds. The social comfort metric, recently proposed by Wang~\cite{wang_junxian_metrics_nodate}, is defined as the  cumulative time where the distance between the robot and the human is within a threshold distance. This is measured for two minimum accepted distances of 0.5 and 1 meters.

\section{Results}
\label{sec:results}


In total we analyze $(N = 18) * 3 = 54$ trials which resulted in $54 * (5$ retrievals$) = 270$ potential encounters between the participants and the robot. 

\subsection{Visualization vs No Visualization}

We compare the baseline \textbf{No AR} condition and our proposed condition (\textbf{Robot+Intent}) in two aspects: Perceived Risk of Collision (\textbf{Q1}) and Perceived Task Efficiency (\textbf{Q2}). For \textbf{Q1} and \textbf{Q2}, a one-way ANOVA test was conducted and significant results were found. A post-hoc t-test between the different trials was then completed, using the Bonferroni method~\cite{pvalueadjust} to adjust the p value by multiplying all raw p values by a factor of $3$ to account for repeated tests. The results are shown in Table~\ref{tab:ar_vs_no_ar}.

\begin{table}[h!]
\centering
\begin{tabular}{|l|c|c|c|}
\hline
 & \textbf{No AR} & \textbf{Robot+Intent} & p value \\ \hline
\textbf{Q1} - Collision Risk & 4.1 ($\sigma$=1.7) & \textbf{2.3} ($\sigma$=1.0) & *0.0002   \\ \hline
\textbf{Q2} - Task Efficiency & 4.4 ($\sigma$=1.8) & \textbf{5.6} ($\sigma$=1.2) & *0.01   \\
\hline
\end{tabular}
\caption{Survey results (7-Likert Scale) for our visualization method vs no visualization. p values are computed for a one-way ANOVA test. * denotes statistical significance}
\label{tab:ar_vs_no_ar}
\end{table}
\vspace{-0.3cm}
These results show that participants felt that our AR-based visualization method significantly improved the perceived safety (affirming \textbf{H1}) and reduced obstruction of the task (affirming \textbf{H2}).

\subsection{Effect of Visualizing the Motion Intent}

We compare the \textbf{Robot Only} condition and our proposed condition (\textbf{Robot+Intent}) for aspects related to perceived safety (\textbf{Q1} and \textbf{Q4}) and perceived task efficiency (\textbf{Q2} and \textbf{Q3}). A t-test is performed for each question. Results are shown in Table~\ref{tab:robot_vs_robotintent}.

\begin{table}[h!]
\centering
\begin{tabular}{|l|c|c|c|}
\hline
 & \textbf{Robot Only} & \textbf{Robot+Intent} & p value \\ \hline
\textbf{Q1} - Collision Risk & 2.8 ($\sigma$=1.1) & \textbf{2.3} ($\sigma$=1.0) & *0.02   \\ \hline
\textbf{Q2} - Task Efficiency & 5.2 ($\sigma$=0.9) & \textbf{5.6} ($\sigma$=1.2) & 0.16   \\ \hline
\textbf{Q3} - Task Completion & 5.7 ($\sigma$=1.1) & \textbf{6.2} ($\sigma$=0.9) & 0.1   \\ \hline
\textbf{Q4} - Collision Prevention & 5.7 ($\sigma$=1.1) & \textbf{6.3} ($\sigma$=0.7) & *0.02   \\
\hline
\end{tabular}
\caption{Survey results (7-Likert Scale) to analyze the effect of adding motion intent visualization on perceived safety and task efficiency. Two-tail p values are computed by t-tests. * denotes statistical significance.}
\label{tab:robot_vs_robotintent}
\end{table}
\vspace{-0.2cm}

Safety: Adding the motion intent visualization to the 3D robot model led to significantly lower perceived safety risk (\textbf{Q1}) and significantly higher sense of collision prevention (\textbf{Q4}). These results affirm \textbf{H3}.

Task efficiency: Adding the motion intent visualization to the 3D robot model tended to increase the survey scores for both questions, \textbf{Q2} and \textbf{Q3}, however statistical significance could not be found. Therefore, \textbf{H4} can not be confirmed with the current data, however, more user studies can provide more information about this hypothesis.

\subsection{Usability}

We compare the \textbf{Robot Only} condition and our proposed condition (\textbf{Robot+Intent}) for aspects related to usability (\textbf{Q5} and \textbf{Q6}). A t-test is performed for each question. Results are shown in Table~\ref{tab:usability}.

\begin{table}[h!]
\centering
\begin{tabular}{|l|c|c|c|}
\hline
 & \textbf{Robot Only} & \textbf{Robot+Intent} & p value \\ \hline
\textbf{Q5} - Easy to Understand & \textbf{6.7} ($\sigma$=0.5) & 6.4 ($\sigma$=0.7) & 0.36   \\ \hline
\textbf{Q6} - Useful Information & 6.1 ($\sigma$=0.7) & \textbf{6.4} ($\sigma$=0.7) & 0.23   \\ \hline
\end{tabular}
\caption{Survey results (7-Likert Scale) to analyze the effect of adding motion intent visualization on usability. Two-tail p values are computed by t-tests. * denotes statistical significance.}
\label{tab:usability}
\end{table}
\vspace{-0.2cm}

We found no statistically significant difference between the two methods. This is because both methods scored fairly high on these two questions (minimum score being 6.1/7.0), especially the question on whether visualizations are easy to understand.

It is interesting to note that \textbf{Q5} is the only survey  question where \textbf{Robot Only} condition scores were higher than the proposed \textbf{Robot+Intent} condition. This shows that there is room for improvement in how to make the information easier to understand to the users via visualization.





\subsection{Task Completion Time}

\vspace{-0.2cm}
\begin{table}[h!]
\centering
\begin{tabular}{|l|c|c|c|}
\hline
 & \textbf{No AR} & \textbf{Robot Only} & \textbf{Robot+Intent}  \\ \hline
 Retrieval Time (s) & \textbf{24.6} ($\sigma$=3.1) & 25.6 ($\sigma$=4.6) & 25.1 ($\sigma$=4.1)   \\ \hline
\end{tabular}
\caption{Mean object retrieval time in seconds across all experiments.}
\label{tab:task_completion_time}
\end{table}
\vspace{-0.2cm}

Mean object retrieval times are shown in Table~\ref{tab:task_completion_time}. \textbf{No AR} condition has the smallest mean and standard deviation in terms of the retrieval time, however, the differences are not significant. Likewise, we did not see any significant differences in the mean retrieval times of different objects.


\subsection{Social Comfort Metric}

The mean time that the robot is within 0.5m and 1m to the human is shown in Table~\ref{tab:comfort_metric}. Interestingly, we find that the \textbf{No AR} condition had the least duration for both versions of the metric, whereas \textbf{Robot+Intent} had the most. One interpretation of these results is that improved user knowledge of the robot's motion makes it more comfortable for the users to be close to the robot.

\begin{table}[h!]
\centering
\begin{tabular}{|l|c|c|c|}
\hline
 & \textbf{No AR} & \textbf{Robot Only} & \textbf{Robot+Intent}  \\ \hline
Comfort Metric (0.5m) & {0.6} ($\sigma$=1.0) & 0.9 ($\sigma$=1.3) & 1.4 ($\sigma$=1.7)   \\ \hline
Comfort Metric (1m) & {4.4} ($\sigma$=3.4) & 5.0 ($\sigma$=3.3) & 6.0 ($\sigma$=4.3)   \\ \hline
\end{tabular}
\caption{Mean comfort~\cite{wang_junxian_metrics_nodate},with two variations, across all experiments.}
\label{tab:comfort_metric}
\end{table}

\vspace{-0.5cm}
\subsection{Discussion}
The user study supports hypotheses \textbf{H1}, \textbf{H2} and \textbf{H3}. Although a trend was present, no significant results were observed for \textbf{H4}, hence further investigation is warranted.

The difference in users' perceived sense of obstruction was shown to only be significant and beneficial in \textbf{Robot+Intent} compared to \textbf{No AR} suggesting that understanding motion intent is important to reducing obstruction within human-robot shared workspaces. One participant (R9) stated that \textit{``even with the visualization of the robot, I still slow down to figure out what the next move is"} for the \textbf{Robot Only} visualization mode, supporting this claim. As well there was the significant difference in the users' perceived ability to prevent collisions (\textbf{Q4}) when comparing \textbf{Robot+Intent} and \textbf{Robot Only}, with \textbf{Robot+Intent} having a greater positive effect. This further supports the claim that understanding motion intent of other agents is important to effectively sharing space.

From the post-experiment feedback form/interview, 12 out of 18 participants mentioned that they found the robot model visualization to be the most helpful visualization system function. This is aided by the higher mean, $6.7$ for \textbf{Robot Only} in Q5 compared to $6.4$ for \textbf{Robot+Intent}. Additionally, Participant (R7) found \textit{``the arrow nice but not super necessary} and while (R10) stated that they \textit{``don't know how necessary it truly is"}, that \textit{``the arrow is too much"} and that the robot model \textit{``is a bit more intuitive"}. This demonstrates that some users did not perceive any benefit from the motion intent visualization with (R7) also stating that more features \textit{``would just start to introduce clutter"} potentially causing \textit{``sensory"} and \textit{``cognitive overload"}. Thus finding a balance between including more visual information and reducing clutter is necessary as was in~\cite{lilija_augmented_2019} and~\cite{gruenefeld_locating_2019}.

Some users noted that some indicator that the robot is close, visual or auditory, would be helpful, with (R9) stating \textit{``add[ing] sound, to warn you that the robot is within some distance"} can be helpful and (R11) suggesting that \textit{``there should be sound to show that you have blindspots"} or \textit{``maybe put an indicator on the right [or left] hand side"}. Further research could include these features to test their efficacy.

\section{Conclusion and Future Work}
\label{sec:conclusion}

We proposed a system that allows users to see mobile robots through walls in AR. The intended use case is to increase the robots' visibility when there is no line of sight between the user and the robot. We validate the effectiveness of visualizing mobile robots in AR with user studies in which a human and a robot have separate tasks, but their paths can cross at a T-junction. Results show that making the robots always visible significantly increased the physical and perceived safety of the users while they felt they could complete the task with less obstructions. 
Results also show that visualizing the robot's movement direction in addition to the robot model further increased the perceived safety of the participants.

There are factors that limit the practical applicability of the proposed system. Our approach requires co-localization of the robot and the AR headset in a common reference frame, and the implemented off-the-shelf localization methods were not robust enough for long, continuous operation. In fact, there were numerous times during the user studies where the localization of the AR headset was lost and had to be re-initialized. Another limitation is the effective field of view of the AR headset, which is smaller than a human's field of view. We observed that users often exhibited ``information seeking" behavior to look for the robot in an effort to compensate for the limited headset field of view. We think that including a method to communicate robot velocity, increasing clarity of robot's future path visualization and increasing the render rate of robot's position and motion intent to at least 60 Hz are potential improvements for a seamless experience and greater adoption of the proposed system.

This paper serves as a proof-of-concept for visualizing occluded entities in AR, and the concept can be further explored. We used a 3D robot model and an arrow for motion intent. Future work to explore other path and motion visualizations would expand this work. Our system presents the AR visualizations regardless of whether the robot is visible to the human or not. The visibility of the robot from the user's perspective can be checked so that only robots under occlusion are presented. How to relay the robot's position through AR when it is not projected in the user's field of view is another interesting research problem. 
It is possible to combine the visualization approach with a human-aware planner to further improve safety. Finally, it would be interesting to extend the work to visualize objects or people that are not visible to the user.  

\section{Acknowledgement}
This project was supported by the Australian Research Council (ARC) Discovery Project Grant DP200102858.

\bibliographystyle{IEEEtran}
\bibliography{refs.bib}
\end{document}